\documentclass{article}
\usepackage{arxiv}

\usepackage{cite}
\usepackage{amsmath,amssymb,amsfonts}
\usepackage{algorithmic}
\usepackage{multirow}
\usepackage{graphicx}
\usepackage{textcomp}
\usepackage{listings}

\usepackage{verbatim} 
\usepackage{soul} 
\usepackage{wasysym} 
\usepackage{todonotes}
\usepackage[utf8]{inputenc}
\usepackage{listings} 

\UseRawInputEncoding

\title{Urdu Speech and Text Based Sentiment Analyzer}

\author{
Waqar Ahmad Jan$^{1}$, Maryam Edalati$^{2}$\\
 \textit{$^{1}$Department of Computer Science and Bioinformatics, Khushal Khan Khattak University, Karak, Pakistan}\\
 \textit{$^{2}$Department of Computer Science, Norwegian University of Science and Technology, Gjøvik, Norway} \\
      waqar.ahmad@kkkuk.edu.pk, maryame@stud.ntnu.no
}

\begin{document}
\maketitle

\begin{abstract}
Discovering what other people think has always been a key aspect of our information-gathering strategy. People can now actively utilize information technology to seek out and comprehend the ideas of others, thanks to the increased availability and popularity of opinion-rich resources such as online review sites and personal blogs. Because of its crucial function in understanding people's opinions, sentiment analysis (SA) is a crucial task. Existing research, on the other hand, is primarily focused on the English language, with just a small amount of study devoted to low-resource languages. For sentiment analysis, this work presented a new multi-class Urdu dataset based on user evaluations. The tweeter website was used to get Urdu dataset. Our proposed dataset includes 10,000 reviews that have been carefully classified into two categories by human experts: positive, negative. The primary purpose of this research is to construct a manually annotated dataset for Urdu sentiment analysis and to establish the baseline result. Five different lexicon- and rule-based algorithms including Naivebayes, Stanza, Textblob, Vader, and Flair are employed and the experimental results show that Flair with an accuracy of 70\% outperforms other tested algorithms.

\end{abstract}


\section{Introduction \& Background}
\label{sec:intro}

An individual's opinion is a judgment, stance, or remark about something \cite{rasheed2018urdu}. Different experts derive different opinions from the same set of evidence. Decisions are heavily influenced by one's point of view. For quick and effective decision-making, people required accurate and authentic information. Human guiding and decision-making rely heavily on people's perspectives, knowledge, and experience \cite{asghar2019creating}. Opinion polarity refers to the view that any entity is either good or bad \cite{bilal2016sentiment}. For example, "Baber Azam is a good Pakistani cricketer" is a positive statement, "He is not a good player" is a negative statement, and "Milk is excellent for health but tea is not" is a neutral statement. 
The origin of the concept, the concept relating entity is described, and the assessment evaluation referred to opinion are the three fundamental aspects of an opinion. The things listed above are required for a recognized viewpoint \cite{daud2015roman}. The viewpoint is gathered by various resources like newspapers, website etc. However, the vast resource of information, views, and concepts is the Internet \cite{asghar2016sentihealth}. Before the invention of the Internet, people gathered information manually, which was time-consuming, labor-intensive, and costly. The internet now makes it easy for us to get a lot of information with just one click, with less time and effort \cite{rasheed2018urdu}. Through internet web users can play a great role in a group work around the world \cite{bilal2016sentiment}. The Internet is a great information medium these days, including social media \cite{kastrati2015analysis}, customer feedback \cite{hao2013visual, edalati2021potential}, job search \cite{miranda2014using}, online shopping \cite{yi2020machine}, commerce \cite{yang2020sentiment}, and MS / PhD research sources \cite{pang2008opinion} based on a literature search.

Providers collect customer ratings for future purposes of improving product quality. The Internet is the birthplace of the modernization of information, but it is not generally seen in an aggregated form and cannot be used directly for decision making. Moreover, it is not possible to manually organize and summarize such large amounts of data. Therefore, effective tools and strategy are needed to extract and summarize opinions.

The research community strives to provide well-organized, systematic and competent data for the knowledge representation. With the development of Web 2.0, posting and assembling views on the web has become much easier. Still, genuine and accurate aggregated data is still a potential research topic. As the opinion score grows, the Natural Language Processing issue creates a new domain related to opinion mining (OM) \cite{bilal2016sentiment}.

\subsection{Opinion Mining}
\label{sec:om}

Opinion Mining (OM) is a new and rising field of research that uses Data Mining (DM) and Natural Language Processing (NLP) approaches to retrieve information and uncover knowledge from text \cite{khan2009mining}.
OM refers to the current trend used in text mining, information retrieval, and computational Linguistics, trying to identify the opinions expressed in natural language text" \cite{khan2015urdu}.
The basic motivation for opinion mining is to extract ideas, opinions and opinions from user suggestions and present the information in an efficient way \cite{batra2021evaluating}. Unlike user experiences for various products, tools and technologies from end users \cite{dalipi2017analysis}, consumers usually present their opinions in the form of review sentences containing a single word or phrase \cite{khan2021urdu}. Text mining, machine learning, and natural language processing techniques are required for proper classification of e-documents \cite{kastrati2019impact, khan2010review}, online news \cite{rana2014news}, instructional videos and resources \cite{kastrati2019integrating, rossi2014language, kastrati2020wet},  blogs \cite{keshtkar2009using}, e-mails, and digital libraries \cite{khan2010sentence}. The function of OM consists of opinion recognition, opinion classification, source recognition, target recognition, and opinion abstraction. View extraction is considered a domain of information identification and total text, following an information retrieval, artificial intelligence, and mathematical device-based approach that distinguishes between observed and actual phrases. Text data can be grouped objectively and subjectively \cite{liu2012survey}. The objective term expresses facts, and the subjective term expresses thoughts, attitudes, or views of any object. Natural language research focuses on extracting real data from the text documents. Subjective analysis related to the analysis of concepts and points from user’s communication and comments. Sentiment analysis and opinion survey are two sub-areas of this field. Some scholars used these terms in reverse although some authors believe that sentiment analysis is a subset of OM. \cite{bakshi2016opinion} indicates OM is almost the same as sentiment analysis. Sentiment analysis is a significant rating of the opinion holder's behavior described in the sentence and can be positive or negative \cite{kastrati2021sentiment}. 

For the opinion mining problem, several approaches have been used, which can be broadly split into two categories: supervised and unsupervised \cite{sadriu2022automated}. The supervised techniques necessitate training data, which necessitates human labour and is usually domain specific. The unsupervised technique is the most widely employed because of two key benefits: There is no need for training data and the domain is independent \cite{khan2013mining}. User Opinion consists of ratings, information gathering, and view ratings. Online data and information on various social networking, commercial, marketing, and corporate websites is usually in an unstructured format. For this reason, research and development in the areas of text mining and opinion mining is very interesting \cite{khan2014mining}.

\subsection{Sentiment Analysis}
Sentiment analysis (SA) is a technique for determining the emotional tone behind a series of words, which is useful for deciphering attitudes, feelings, and opinions expressed in an online forum \cite{imran2020cross, shariq2020cross, khan2020urdu}. Fine-Grained Sentiment Analysis, Emotion Detection, Aspect-Based Sentiment Analysis, and Multilingual Sentiment Analysis are some of the different methods of sentiment analysis \cite{kastrati2020weakly, abdelwahab2017enhancing, 10.1145/3404555.3404633}. Sentiment analysis has been approached as a natural language processing (NLP) problem, with three levels of execution: document/snippet level, sentence level, and aspect level \cite{rehman2016lexicon}. The majority of sentiment analysis research has focused on approaches like Naive Bayesian, decision tree, support vector machine, and maximum entropy \cite{maghilnan2017sentiment}. Languages such as English, Arabic, Italian, Chinese, and a few other widely used languages have been the focus of earlier research in this area \cite{khan2020urdu}. As a result, the unimportant or deceptive phrases are ignored by the polarity classifier. Because gathering and labelling information at the sentence level takes time and is difficult to test. The text is rated as positive or negative, for example, thumbs up or thumbs down. Some classifiers employ a multi-point range, such as a five-star rating system for movie reviews \cite{syed2010lexicon}. Positive and negative opinions conveyed in the document were classified using sentiment polarity analysis \cite{khan2010sentence}. A sentiment analyzer's main purpose is to classify a person's subjectivity orientation as positive or negative. As a result, sentiment analysis is also known as subjectivity analysis. These techniques can be divided into two categories. The first is the machine learning method. A machine learning strategy can be supervised (for example, classification that uses labelled data to train the classifiers) or unsupervised (for example, classification that utilizes unlabeled data to train the classifiers) (e.g. clustering). The semantic orientation (SO) strategy, also known as the lexicon-based approach, requires no prior trained data and merely considers the positive or negative orientation of words \cite{mukhtar2020effective}. With the growing popularity of the Internet, Urdu websites, like those for other languages, are getting increasingly popular as people prefer to share their thoughts and feelings in their own language. That is why, before casting a vote or purchasing the latest gadget, we look for other people's comments and reviews on the Internet \cite{syed2010lexicon}. 

Sentiment evaluation is utilized in numerous applications. Here, we use it to understand the mindset of people based mostly on their conversations with every other. For a machine to know the mindset/mood of the people by way of a conversation, it wants to know who're interacting within the dialog and what is spoken so we implement a speaker and speech recognition system initially and carry out sentiment evaluation on the information extracted from prior processes. Understanding the temper of people might be very helpful in many instances. For example, computer systems that possess the ability to understand and reply to human non-lexical communication such as emotions. In such a case, later detecting humans’ emotions, the machine may customize the settings according his/her wants and preferences \cite{asghar2019creating}. 

The strategy adopted within the notes investigates the challenges’ and strategies to carry out audio speech sentiment evaluation on audio recordings utilizing speech recognition. We use speech recognition libraries to transcribe the audio recordings and a proposed speaker discrimination method based on sure speculation to determine the audio system concerned in a conversation. Further, sentiment evaluation is carried out on the speaker particular speech knowledge which permits the machine to understand what the people have been speaking about and the way they feel.
In our proposed system we used Naive Bayes, Textblob, Stanza, Flair, and Vader strategies to perform sentiment analysis and compared them to find the most environmental friendly algorithm for our needs.

\subsection{Speech Recognition}
\label{sec:sr}
Speech recognition refers to the ability of a machine or program to recognize words and phrases in spoken language and convert them to a machine-readable format that can then be processed \cite{sabzi2019comparative}. We employed speech recognition tools like Google Speech Recognition in this work. A comparison is performed, and the ideal suite for the suggested model is selected.
Speech recognition software breaks down a speech recording's audio into individual sounds, analyses each sound, employs algorithms to find the most likely word slot in that language, and then transcribes those sounds into text. Natural Language Processing (NLP) and deep learning neural networks are used in the speech recognition computer code. According to the Algorithm magazine, "NLP may provide a way for computers to examine, understand, and derive meaning from human language in a highly sensible and beneficial way." This implies that the computer code breaks down the speech into bits that it can understand, converts it to a digital representation, and examines the content elements. The computer code then makes decisions based on supported programming and speech patterns, forming hypotheses about the user's true vocal communication. The computer algorithm transcribes the language into text after determining what the users may have spoken. This all seems simple enough, but because to technological advancements, these many, intricate procedures are happening at breakneck speed. Human speech will be transcribed much more reliably, correctly and swiftly by machines than by humans.

\subsection{Evaluating Expression}
\label{sec:ee}

Various strategies were used to extract opinion elements in text documents or evaluating statements and feedback \cite{edalati2021potential}. Several other techniques are also used, such as supervised, unsupervised and semi-supervised methods. The supervised learning method uses a machine learning strategy prepared by manually identified information to acquire and classify the attribute set identified in the evaluation \cite{bilal2016sentiment}.  
Although supervised machine learning produces more accurate results than lexicon-based approaches, the former takes too long because the classifier must first be taught. As a result, when dealing with large amounts of data, this method is ineffective. Second, supervised machine learning relies on training data to work \cite{khan2022multi}. 
The unsupervised learning approach does not require the independence of trained and manually tagged data and its domain. Semi-supervised learning techniques, on the other hand, consist of pairs of marked and unmarked information \cite{othman2017extracting}.

\subsection{Urdu Language}
\label{sec:ul}
Approximately 60.5 million speakers, mainly in the Indian subcontinent \cite{hussain2006urdu}. The Indo-Pak subcontinent is a major market for a wide variety of goods. People who speak Urdu or Hindi respond to any online product or event using text written in Roman Urdu, Roman Hindi, pure Urdu, or pure Hindi. Hindi and Urdu are the most widely spoken languages in the Indian subcontinent. Urdu and Hindi are spoken by more than 588 million people, which is more than the number of people who speak English \cite{khan2021review}.
Urdu is the widely spoken Indo-Aryan. There are numerous speakers in Pakistan, India, Afghanistan, Iran and Bangladesh. In addition, Urdu is Pakistan's official language. Urdu orthography is similar to Arabic, Persian, and Turkish. Italic Arabic and Nastaliq style are used \cite{riaz2007challenges}. However, Urdu is still considered a low-resource language due to lack of tokenizers, stemming, and large publicly available corpora. So, training models for Urdu language remains a challenge. Researchers now address this issue by generating synthetic samples using text generation techniques \cite{fatima2022systematic}. 

\textbf{Urdu vs English}: Sentiment analysis of English text is well studied. In \cite{khan2015urdu} and \cite{annett2008comparison} a very extensive and comprehensive survey is presented. The core approach for processing English texts (machine learning, detailed in Related Work) can be used for Urdu texts, while modifications and adjustments are orthography, morphology, and grammar between the two.  

\textbf{Urdu vs Arabic}: The main language comparable to Urdu is Arabic. In the field of computational linguistics, Arabic is much more mature than Urdu, and several approaches to Arabic word processing have been proposed. Orthodox and morphologically, the two languages are very similar, but Urdu grammar tends towards Sanskrit and Persian \cite{riaz2007challenges}.

\textbf{Urdu vs Hindi}: Hindi is the main dialect of Urdu, with minimal differences in the grammar of the two languages. However, their spelling and vocabulary are different. Urdu, from right to left, uses the Persian-Arabic Nastaliak calligraphy style and uses the vocabulary derived from Arabic and Persian. Hindi, on the other hand, uses Devanagari from left to right to derive vocabulary from Sanskrit.

\section{Related Work}
\label{sec:relatedwork}

Text classification has become one of the most important strategies for handling and organizing text data, because of the growing expansion of online information \cite{shaikh2021towards}. Text categorization algorithms are used to classify news stories, identify interesting material on the World Wide Web, and lead a user's hypertext search. Text categorization begins with the transformation of documents, which are often strings of characters, into a representation suited for the learning algorithm and the classification task \cite{rasheed2018urdu}. 

The most important aspect of opinion mining is extracting and analyzing people's feedback in order to learn about their feelings. New opportunities and difficulties have arisen as the availability of opinion-rich resources such as internet blogs, social media, and review sites has increased \cite{pang2008opinion}. People are now using the internet for social relations, business correspondence, e-marketing, e-commerce, and e-surveys, among other things, thanks to widespread use of computers, cellphones, and high-speed internet. People discuss a product, service, political entity, or current events by sharing their ideas, suggestions, comments, and opinions \cite{bilal2016sentiment}. Positive, negative, and neutral comments are the three types of sentiment. Positive comments are those where the consumer praises a particular feature of the product. Negative comments are the ones in which a person expresses dissatisfaction with a product. While neutral comments are those in which a user simply replicates some of the product's features or simply responds to other users' comments with no polarity \cite{daud2015roman}. 

The research work in \cite{rasheed2018urdu} used datasets containing significant details that can be used to make frequent decisions. It is difficult for any system to make intelligent decisions with no datasets available. As a result, classification algorithms simplify the work by isolating useful models and emphasizing key data categories. All relevant documents can be divided into three categories: supervised, unsupervised, and semi-supervised. SVM, ANN, NB, KNN, and Decision Trees are some of the approaches used to categorise texts. In the current investigation, however, only SVM, J48, and KNN were used to evaluate the data. Same work on dataset can also be done using Naïve Bayes, NLTK Vader, Stanza, Flair and Textblob classifiers .

Zubair et al., in \cite{asghar2019creating} gathered reviews from a variety of sources. There are 493 drug reviews (DR), of which 55\% are positive and 45\% negative. The mobile reviews (MR) dataset comprises 373, with a ratio of 58\% positive reviews, 42\% negative reviews. In the book domain (BR), there are 431 reviews, with 65\% positive and 35\% negative. In the mixed dataset, there are 1,201 reviews. To assemble the testing and training corpora, the reviews are saved in two different text files. To determine the sentiment of sentences in datasets of positive and negative sentences, used a proposed Urdu lexicon to do sentiment categorization. 

The researchers in \cite{bilal2016sentiment} used five steps in the suggested model. First, using Easy Web Extractor software, opinions written in Roman-Urdu are extracted from a blog. The retrieved opinions are documented in text files to create a training dataset with labelled instances of 150 positive and 150 negative opinions. Using WEKA's Text Directory Loader command in Simple CLI mode, the dataset is first transformed to ARFF (Attribute-Relation File Format). The ARFF dataset is then placed into the WEKA explorer mode as a training data set for the machine to learn from. The data is initially preprocessed using WEKA filters, and then three distinct algorithms are applied to the dataset to train the machine and generate three models: Nave Baysian, KNN, and Decision Tree. The three models are given a testing data set, and the results are analyzed in each case \cite{bilal2016sentiment}. Similar study is presented in \cite{chandio2022attention}, where the authors incorporated attention-based mechanism for identifying polarity in Roman Urdu.

The study conducted in \cite{daud2015roman} aimed to give non-Urdu speaking clients the ability to benefit from comments written in Roman Urdu. A website called whatmobile\footnote{https://www.whatmobile.com.pk/} has been chosen for this experiment. This website provides information about popular cell phone brands in Pakistan. This site also allows users to leave comments on each cell phone page. The majority of the comments are written in Roman Urdu, however some individuals also provide English remarks. There are four main steps involved in the system. Crawling the web, translating Roman-Urdu reviews into English, determining the polarity of opinions, and assigning a graphical rating. The review miner will crawl all of the reviews from the input URL and save them to a local temporary storage location. The data will be passed to the Review Translation module after it has been retrieved. The Review Translator module will use the Microsoft Bing Translation Service API to query the comments. The next step is to analyse and tag the translated text once it has been received.

The research work in \cite{asghar2016sentihealth}, focuses on some of the relevant studies on the construction of sentiment lexicons in general purpose and domain specialized paradigms utilizing boot-strapping and corpus-based strategies. The project aimed to create a sentiment lexicon to improve the efficiency of health-related sentiment analysis applications by resolving issues such as low coverage of health-related content in existing lexicons, such as SWN, incorrect sentiment class assignment to domain specific words, and inaccurate scoring of health-related words. 
The data collection and preprocessing module aims to collect data from a variety of sources, including online health forums and publicly available datasets, and to remove noise from the collected data using various preprocessing steps, including tokenization, stop word removal, lemmatization, spell correction, and co reference resolution. 

The authors in \cite{khan2015urdu} used preprocessing and feature extraction in MATLAB, whereas classification is done using J-48 algorithms. Because there is currently no reference database for Urdu characters, one is being constructed that includes both handwritten and machine-written scanned images. There are 441 Urdu language characters in the database.  The three primary steps including preprocessing, feature extraction, and classification were used for experimental analysis approach of OCR.

The study conducted in \cite{khan2021urdu} focuses on Urdu that is still in its early stages. Support vector machine (SVM), Naive Bayes (NB), random forest (RF), AdaBoost, multilayer perceptron (MLP), logistic regression (LR), 1-dimensional convolutional neural network (1D-CNN), and long short-term memory machine learning and deep learning models (LSTM) are utilized for effectiveness and state of the art results. All of these machine and deep learning models have been deployed in their proposed UCSA corpus .

\section{Material \& Methods}
\label{sec:mm}
With the explosive growth of social media (e.g., reviews, forum discussions, blogs, micro-blogs, Twitter, comments, and postings in social network sites) on the Web, individuals and organizations are increasingly using the content in these media for decision making \cite{imran2020cross}. Nowadays, if one wants to buy a consumer product, one is no longer limited to asking one’s friends and family for opinions because there are many user reviews and discussions in public forums on the Web about the product.
It's not easy to create an opinion analysis for the Urdu (Voice, Text) sentiment analysis without rich understanding of data science and natural language processing \cite{chandio2022attention}. The preparatory processes are carried out using Python 3.7 and the IDLE, IDE as well as the relevant libraries mentioned in this section. 

This section provides an overview of existing supervised and unsupervised algorithms that have been widely used to extract targets from sentence datasets. The main goal of this research work is to identify potential ways for extracting opinion targets that could be beneficial in opinion mining. The investigation of the factors that influence the current supervised, unsupervised learning approach of opinion target extraction is thus the article's key contribution.

\subsection{Tool \& Libraries}
\label{sec:tl}

\textbf{Python}: Python is a high-level, general-purpose programming language that is widely used. The Python programming language (the most recent version is Python 3.10.1) is utilized in web development, machine learning applications, and every other cutting-edge technology in the software industry.

\textbf{IDE}: An integrated development environment (IDE) is a software package that includes a code editor, build automation, tools, and a debugger. In addition to learning and educational purposes, IDLE offers the core packages and concepts of an IDE.

\textbf{Jupyter Notebook}
Jupyter Notebook is an open source web tool for creating and sharing documents with live code, equations, visualisations, and text. Project Jupyter is in charge of maintaining Jupyter Notebook.

\textbf{Libraries to import}: 
\begin{enumerate}
    \item import pandas
    \item import numpy
    \item import flair
    \item import TextBlob
    \item import speech$\_$recognition
    \item import translator 
    \item import stanza
    \item import nltk
\end{enumerate}

\textbf{Tokenization and Sentence Split}:
 As the initial stage in processing, Stanza tokenizes raw text and groups tokens into sentences. Stanza, unlike other existing toolkits, integrates raw text tokenization and sentence segmentation into a single module. This is done to forecast the position of words in a sentence because the use of words in some languages is context-sensitive.
 
\textbf{Multi-Word Token Expansion}: The algorithms described above identify multi-word tokens, which are then expanded into syntactic words as the basis for downstream processing. This is accomplished by employing a sequence-to-sequence (seq2seq) model to ensure that expansions in the training set are commonly observed, as they are always robustly expanded, while retaining the flexibility to statistically predict unknown words.

\textbf{POS and Morphological Feature Tagging}: Stanza allocates each word in a phrase to a part-of-speech (POS) and analyses its universal morphological features (UFeats, such as singular/plural, 1st/2nd/3rd person, and others) before tagging it.

\textbf{Lemmatization}: Stanza additionally lemmatizes each word in a phrase to return it to its original form (for example, did-->do). Stanza's lemmatizer is an ensemble of a dictionary-based lemmatizer and a neural seq2seq lemmatizer, similar to the multi-word token expander. In addition, using the encoder output of the seq2seq model, an extra classifier is developed to predict shortcuts like lowercasing and identity copy for resilience on long input sequences like URLs.
Stanza is free and open-source, with pre-trained models accessible for all supported languages and datasets. Stanza, according to the researchers, will enable multilingual NLP research and applications \cite{ghafoor2021impact}, as well as urging new study into a wide spectrum of human languages.

To begin with, the Stanza models that can be downloaded have only been trained on a single dataset. To test its resilience, they must train the models with data gathered from a variety of sources. Second, the library is tuned for precision, which might sometimes come at the expense of computational efficiency, restricting the toolkit's application. Finally, for richer text analytics, the researcher will need to make it compatible with various NLP techniques, such as neural co-reference resolution or relation extraction.

\subsection{Dataset}
\label{sec:dataset}

We used a publicly available large scale Urdu Tweet dataset containing almost 1140825 tweets consisting of short and long characters messages from 23 to 24 (short) and up to 280 (long) tweets with different emojis (721) \cite{batra2021large}. This dataset have been collected of two months September, October 2020 using streaming API (live data) and search API (historical data) using BirdIQ (search API) after getting API Key and API secret from Tweeter for authentication. The whole dataset comprises four columns (ID, Text, Emoticon and Category). The ID was further filtered for duplication. To get polarity (+ve, -ve and neutral) the dataset was categorized by separating different corpus, emojis from the text messages.

\section{Sentiment Analyzer Tool}
\label{sec:results}
This section discusses the findings of the experiments, as well as how they relate to the goals of the study. To test the performance of five algorithms, five lexicon-based models were utilized. When applied to the dataset, the algorithm with the best accuracy is selected.

Generally, SA models can be modeled using unsupervised approach or supervised approach \cite{Kia2016Multilingual}.

Unsupervised techniques do not require marked or labeled data, and they automatically select target features based on semantic relatedness and syntactic patterns.
This proposed system is a multi-layered polarity identification model for analyzing sentiments of any piece of voice message in Urdu and suggests its polarity as positive, negative or neutral. The Google API 'Application Programmable Interface', allows you to systematically incorporate your product with Google and encompasses a speech recognition API to converts spoken text (microphone) into written communication, in brief Speech to Text. You’ll merely speak in a very electro-acoustic transducer and Google API can translate this into written communication.

\begin{figure}[ht]
    \centering
    \includegraphics[width=16cm]{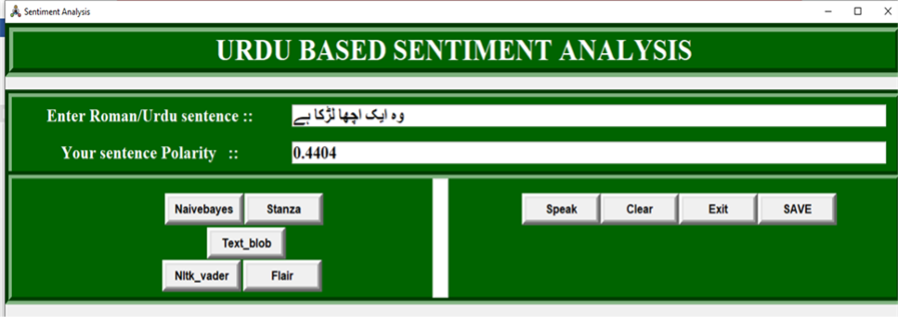}
    \caption{Proposed Sentiment Analysis Tool}
    \label{fig1}
\end{figure}

To get sentiment analysis polarity, different libraries for classifiers like “textblob”, “Stanza”, “Textblob”, “NLTK Vader” and “Flair” are required to be installed in Python. 

\subsection{Speak Button}
 
One of the most essential jobs in the domain of human-computer interaction is speech recognition. The term "speech recognition" refers to the automatic recognition of human speech. Speech recognition has a wide range of uses, from automatic transcription of speech data (such as voicemails) to interfacing with robots through speech.

Speech is, of course, the first component of speech recognition using phonetic and articulatory features \cite{shahrebabaki2019phonetic, local2012phonetic, ali2014dwt}. With a microphone, actual sound is turned to an electrical signal, which is then converted to digital data with an analog-to-digital converter. After the audio has been digitized, numerous models can be employed to convert it to text then analyze accordingly \cite{ali2019regularized, shahrebabaki2018acoustic}.

In our proposed model, when speak button is pressed, the system is waiting for 4 seconds to user for input/record his voice message, and the $recognizer()$ class is to recognize voice message or speech.
Using $recognize\_google()$, Google Web Speech API is free methods and does’t require API key for recognizing speech from an audio source. Others below recognizing method can also be used by $recognizer()$ class

\begin{lstlisting}
    recognize_bing() --> function uses Microsoft’s cognitive services.
    recognize_google() --> function uses Google’s free web search API.
    recognize_google_cloud() --> function uses Google’s cloud speech API
    recognize_wit() --> function uses the wit.ai platform.
\end{lstlisting}

All these APIs require authentication via an API key or a username/password combination.
You can simply install using “pip install SpeechRecognition” in Gitbash terminal. 
By calling the recognizer class, we assign ‘r’ an instance to a variable.

\begin{lstlisting}
    r = sr.Recognizer()
\end{lstlisting}

When working with an audio file, the following threshold method will increase speech recognition.

\begin{lstlisting}
    r.energy$_$threshold=300
\end{lstlisting}

\subsection{Record Method}
The record function will record audio data from the beginning of the file until it runs out. However, we can change this by assigning values ‘timeout=4’ to them. Similarly ‘duration’ and ‘offset’ can also be used for changing recording method values time limits in seconds.

\begin{lstlisting}
    with sr.Microphone() as source:
            audio = r.listen(source, timeout=4)
            try:
                self.first.set(r.recognize_google(audio,language='ur-PK'))
            except sr.UnknownValueError:
                messagebox.showerror("Error","Could't recognize your voice)
\end{lstlisting}

\subsection{Clear Button}
The following code is invoked to clear all the textboxes whenever clear button is pressed:

\begin{lstlisting}
    def c(self):
        self.first.set("")
        self.second.set("")
\end{lstlisting}        

\subsection{Exit Button}
The exit button is used to close whole window appearing prior dialogue box to ask for (yes or no), so when pressed ‘yes’ the whole window will become close and the program become ready for recompilation although when pressed ‘No’ the whole compiled results will remain unchanged.

\begin{lstlisting}
    def exit(self):
        op = messagebox.askyesno("Exit", "DO YOU WANT TO EXIT")
        if op > 0:
             self.root.destroy()
\end{lstlisting}

\subsection{Save Button}
This button is used to make a CSV (Comma Separated Value) file and save all the data, after getting polarity results of a sentence exists in second entrybox along with sentence in first entrybox. The necessary code is appended below:

\begin{lstlisting}
    def sa(self):   
        with open('allsentiment5.txt','a',encoding='UTF-8',newline='') as csvfile:
        fieldname = ['Sentence','Naive bayes','Textblob',
        'Flair', 'Nltk-Vader ',' 'Stanza']
        csv_writer=DictWriter(csvfile,fieldnames=fieldname, delimiter='\t')
        csv_writer.writerow({
        'Sentence':self.first.get(),
        'Naive bayes':polarity2,
        'Textblob':polarity,
        'Flair':sentence1.labels,
        'Nltk-Vader':ss['compound'],
        'Stanza':sentence.sentiment})
\end{lstlisting}

\subsection{Naive Bayes Button}
It is based on the concept of conditional probability. The probability that something will happen if something else has already happened is known as conditional probability. We can calculate the likelihood of an event using conditional probability and prior knowledge. According to Bayes theorem:

$Posterior = likelihood * proposition/evidence $

In our model the following python code will calculate polarity of a sentence whether it is positive or negative:

\begin{lstlisting}
    translator = Translator()
    result2 = translator.translate(text=self.first.get(), src='ur', dest='en')
\end{lstlisting}

By default, $Translator.translate$ detects the language of the supplied text in sentence entrybox which then be converted from source (Urdu) to its destination (English).

\begin{lstlisting}
    blob = TextBlob(result2.text, analyzer=NaiveBayesAnalyzer())
\end{lstlisting}

TextBlob is a text processing package for Python 2 and 3. It offers a basic API for doing standard natural language processing (NLP) activities like part-of-speech tagging, noun phrase extraction, sentiment analysis, classification, and translation, among others.
PatternAnalyzer (based on the pattern library) and NaiveBayesAnalyzer (an NLTK classifier trained on a movie reviews corpus) are the two sentiment analysis implementations in the $textblob.sentiment$.

\begin{lstlisting}
    print (blob.sentiment.classification)
\end{lstlisting}

This will print the result of the NaiveBayesAnalyzer in namedtuple form Sentiment (classification, p\_pos, p\_neg).

\subsection{Stanza Button}

Stanza is a Python library for natural language processing. This button utilized tools for converting a string of human language text into lists of sentences and words, generating basic forms of those words, parts of speech, and morphological features, performing a syntactic structure dependency parse, and recognizing named entities in a pipeline. Stanza is made up of highly accurate neural network components that allow for quick training and evaluation using annotated data. The PyTorch library is used to build the modules.

Simply write below commands in gitbash terminal to install stanza library.

\begin{lstlisting}
    pip install googletrans==3.1.0a0
    pip install stanza
\end{lstlisting}

It’s as simple as downloading a pre-trained model and setting up a pipeline. 

\begin{lstlisting}
    stanza.download('urdu')
    stanza.download('english')
\end{lstlisting}

Specifying the processors:

\begin{lstlisting}
    translator =Translator()
        value=translator.translate(sentance, dest='en',src='ur')
        trn=value.pronunciation
\end{lstlisting}

By default, $Translator.translate$ detects the language of the text entered in the sentence entrybox and converts it from Urdu to English (English).

\begin{lstlisting}
    nlp = stanza.Pipeline(lang='en', processors='tokenize,sentiment')
    doc = nlp(trn)
        for i, sentence in enumerate(doc.sentences):
		polarity= sentence.sentiment
            if(polarity==0):
                self.second.set((polarity,':-1 Negative'))
            elif(polarity==1):
                self.second.set((polarity,':0 Neutral'))
            elif(polarity==2):
                self.second.set((polarity,':1 positive'))
\end{lstlisting}

The $TokenizeProcessor$ is usually the pipeline's initial processor. At the same time, it does tokenization and sentence segmentation. This processor divides the raw input text into tokens and sentences, allowing for sentence-level annotation later on. The name tokenize can be used to call this processor depending on the use-case one might need to specify a set of processors and the package to fetch the different processors from.

\subsection{TextBlob Button}
TextBlob is a Python package offering a simple API for interacting with its functions and perform basic NLP tasks.

By default, $Translator.translate$ recognizes the language of the text entered in the sentence entrybox, which subsequently be transformed from Urdu to English using the following commands:

\begin{lstlisting}
    translator=Translator()
    result=translator.translate(text=self.first.get(),src='ur',dest='en')
\end{lstlisting}

The polarity and subjectivity of a statement are returned by TextBlob button. The range of polarity is in [-1, 1], with -1 indicating a negative sentiment and 1 indicating a positive sentiment. Negative words are used to change the polarity of a sentence. The position of personal opinion and factual information in a text is measured by subjectivity. Because of the text's highest subjectivity, it contains personal opinion rather than factual information. For example, we determined the polarity and subjectivity of the text given in Figure \ref{fig1}:

\begin{figure}[ht]
    \centering
    \includegraphics[width=12cm]{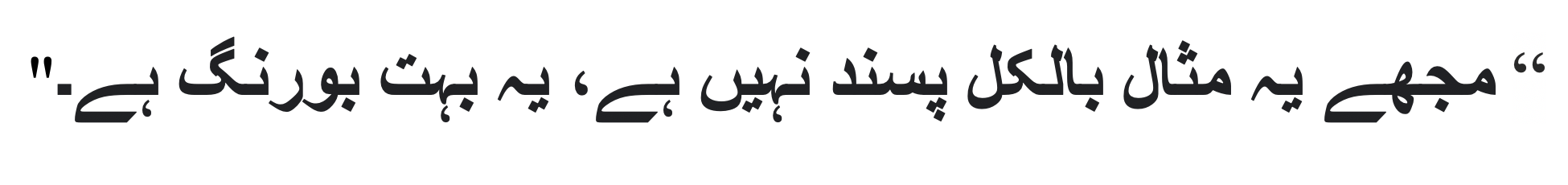}
    \caption{example 1}
    \label{fig1}
\end{figure}

\begin{figure}[ht]
    \centering
    \includegraphics[width=12cm]{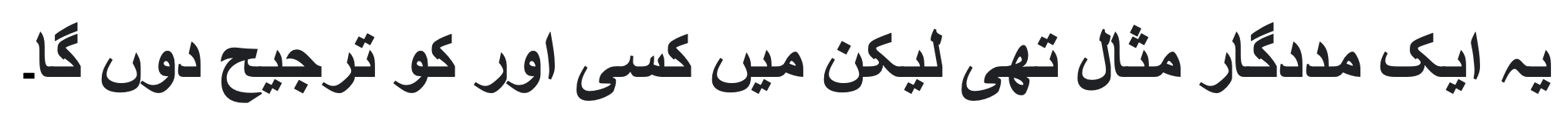}
    \caption{example 2}
    \label{fig2}
\end{figure}

The sentence "I do not like this example at all, it is too boring” has a polarity -1 and subjectivity 1, which is reasonable. However, the sentence shown in Figure \ref{fig2}, 

with its English version "This was a helpful example but I would prefer another one”, gives us a 0.0 for both subjectivity and polarity, which isn't the result we would expect to get. To get polarity the following code can be used:

\begin{lstlisting}
    polarity=TextBlob(result.text).sentiment.polarity
    self.second.set(polarity)
\end{lstlisting}

\subsection{NLTK Vader Button}
VADER (Valence Aware Dictionary for Sentiment Reasoning) is a text sentiment analysis model that takes into account both the polarity (positive/negative) and the intensity (strength) of emotion or we can say that Vader is a natural language processing (NLP) system that combines a sentiment lexicon approach with grammatical rules and syntactical conventions to represent sentiment polarity and intensity. It's included in the NLTK package and may be applied on unlabeled text data.
The language of the text entered in the sentence entrybox is recognised by $Translator.translate$, which is then converted from Urdu to English using the following commands:

\begin{lstlisting}
    translator = Translator()
    result1 = translator.translate(text=self.first.get(), src='ur', dest='en')
  \end{lstlisting}      
        
Create a SentimentIntensityAnalyzer object.

\begin{lstlisting}
    sid = SentimentIntensityAnalyzer()
\end{lstlisting}  

The following code is to get $polarity\_scores$ method of $SentimentIntensityAnalyzer$ object gives a sentiment dictionary which contains pos, neg, neu, and compound scores.

\begin{lstlisting}
    ss = sid.polarity_scores(result1.text)
    self.second.set(ss['compound'])
\end{lstlisting}  

\subsection{Flair Button}

Zalando Research developed and open-sourced Flair, a simple natural language processing (NLP) library. The Flair framework is based on PyTorch. Flair uses a pre-trained model to detect positive or negative remarks and prints a prediction confidence value in brackets behind the label.

\begin{lstlisting}
    sentiment_model = flair.models.TextClassifier.load('en-sentiment')
\end{lstlisting}  
  
The global variable “sentence1” is used to access this variable inside as well as outside of the function while $translator.translate$ recognises is the language of the text given in the sentence entrybox, which is then transformed from Urdu to English using the following commands:

\begin{lstlisting}
    global sentence1
        translator = Translator()
        result = translator.translate(text=self.first.get(),
        src='ur', dest='en')
    Text Sentiment analysis

    sentence1 = flair.data.Sentence(result.text)
    sentiment_model.predict(sentence1)
    self.second.set(sentence1.labels)
\end{lstlisting} 

The tool currently only support lexicon-based models. However, there have been many recent studies where sentiment analysis is conducted employing deep learning models \cite{naqvi2021utsa, ghulam2019deep}. Although for Urdu language due to limited availability of publicly available data sets, the need for synthetic text generation \cite{zhu2018texygen} for SA model training, as in \cite{imran2022impact} is necessary for providing deep learning model's support. This research work is however limited in this respect. Furthermore, semantics \cite{kastrati2015semcon, saeed2021enriching, kastrati2016semcon} and concept space \cite{zhou2018deep, kastrati2019performance}, as well as ontology models \cite{rajput2014ontology, kastrati2015improved, nargis2016generating}, and processing systems \cite{thaker2015domain} can be utilized to enrich Urdu lexicon for developing polarity assessment models for Urdu \cite{mukhtar2018lexicon}, rather than using translation services to convert text into English first. 

\section{Results}
\label{sec:results}
We used the lexicon based prediction models as explained above on the dataset mentioned in section 3.2. To test the performance of the models on the dataset, we manually labelled 10,000 samples from the Urdu Tweet dataset into positive and negative labels. 

The results are depicted in following tables for various models. 

Table 1 depicts the confusion matrix for Naive Bayes classifier.

\begin{table}[htb!]
\centering
\caption{Naive Bayes Confusion Matrix}
\label{tab:tab1}
\begin{tabular}{|l|ccc|}
\hline
\multirow{2}{*}{}      & \multicolumn{3}{c|}{Predicted Values}                                                         \\ \cline{2-4} 
                       & \multicolumn{1}{l|}{}         & \multicolumn{1}{l|}{Positive} & \multicolumn{1}{l|}{Negative} \\ \hline
\multicolumn{1}{|c|}{\multirow{2}{*}{Actual Values}} & \multicolumn{1}{c|}{Positive} & \multicolumn{1}{c|}{4860} & 1721 \\ \cline{2-4} 
\multicolumn{1}{|c|}{} & \multicolumn{1}{c|}{Negative} & \multicolumn{1}{c|}{2406}     & 1013                          \\ \hline
\end{tabular}
\end{table}

The accuracy achieved is 59\%.

The results obtained by Textblob classifier are depicted in Table 2. 

\begin{table}[htb!]
\centering
\caption{Textblob Confusion Matrix}
\label{tab:tab2}
\begin{tabular}{|l|ccc|}
\hline
\multirow{2}{*}{}      & \multicolumn{3}{c|}{Predicted Values}                                                         \\ \cline{2-4} 
                       & \multicolumn{1}{l|}{}         & \multicolumn{1}{l|}{Positive} & \multicolumn{1}{l|}{Negative} \\ \hline
\multicolumn{1}{|c|}{\multirow{2}{*}{Actual Values}} & \multicolumn{1}{c|}{Positive} & \multicolumn{1}{c|}{4862} & 1719 \\ \cline{2-4} 
\multicolumn{1}{|c|}{} & \multicolumn{1}{c|}{Negative} & \multicolumn{1}{c|}{2404}     & 1015                          \\ \hline
\end{tabular}
\end{table}

The accuracy achieved is also 59\%.

Flair results are depicted in Table 3.

\begin{table}[htb!]
\centering
\caption{Flair Confusion Matrix}
\label{tab:tab3}
\begin{tabular}{|l|ccc|}
\hline
\multirow{2}{*}{}      & \multicolumn{3}{c|}{Predicted Values}                                                         \\ \cline{2-4} 
                       & \multicolumn{1}{l|}{}         & \multicolumn{1}{l|}{Positive} & \multicolumn{1}{l|}{Negative} \\ \hline
\multicolumn{1}{|c|}{\multirow{2}{*}{Actual Values}} & \multicolumn{1}{c|}{Positive} & \multicolumn{1}{c|}{5390} & 1191 \\ \cline{2-4} 
\multicolumn{1}{|c|}{} & \multicolumn{1}{c|}{Negative} & \multicolumn{1}{c|}{1822}     & 1597                          \\ \hline
\end{tabular}
\end{table}

The accuracy achieved is 70\%.

Confusion matrix for Vader is depicted in Table 4.

\begin{table}[htb!]
\centering
\caption{Vader Confusion Matrix}
\label{tab:tab4}
\begin{tabular}{|l|ccc|}
\hline
\multirow{2}{*}{}      & \multicolumn{3}{c|}{Predicted Values}                                                         \\ \cline{2-4} 
                       & \multicolumn{1}{l|}{}         & \multicolumn{1}{l|}{Positive} & \multicolumn{1}{l|}{Negative} \\ \hline
\multicolumn{1}{|c|}{\multirow{2}{*}{Actual Values}} & \multicolumn{1}{c|}{Positive} & \multicolumn{1}{c|}{3640} & 698 \\ \cline{2-4} 
\multicolumn{1}{|c|}{} & \multicolumn{1}{c|}{Negative} & \multicolumn{1}{c|}{1444}     & 852                          \\ \hline
\end{tabular}
\end{table}

The accuracy is 45\%.

With Stanza the accuracy is only 31\% and the confusion matrix is depicted in Table 5. 

\begin{table}[htb!]
\centering
\caption{Stanza Confusion Matrix}
\label{tab:tab5}
\begin{tabular}{|l|ccc|}
\hline
\multirow{2}{*}{}      & \multicolumn{3}{c|}{Predicted Values}                                                         \\ \cline{2-4} 
                       & \multicolumn{1}{l|}{}         & \multicolumn{1}{l|}{Positive} & \multicolumn{1}{l|}{Negative} \\ \hline
\multicolumn{1}{|c|}{\multirow{2}{*}{Actual Values}} & \multicolumn{1}{c|}{Positive} & \multicolumn{1}{c|}{1408} & 2035 \\ \cline{2-4} 
\multicolumn{1}{|c|}{} & \multicolumn{1}{c|}{Negative} & \multicolumn{1}{c|}{393}     & 1656                          \\ \hline
\end{tabular}
\end{table}

The accuracy of 31\% achieved by Stanza on 10,000 samples is considered low compare to other classifiers like Textbob, Naivebayes, Flair and NLTK\_Vader. Flair performs the best in our case with an accuracy of 70\%. 

\begin{figure}[ht]
    \centering
    \includegraphics[width=16.5cm]{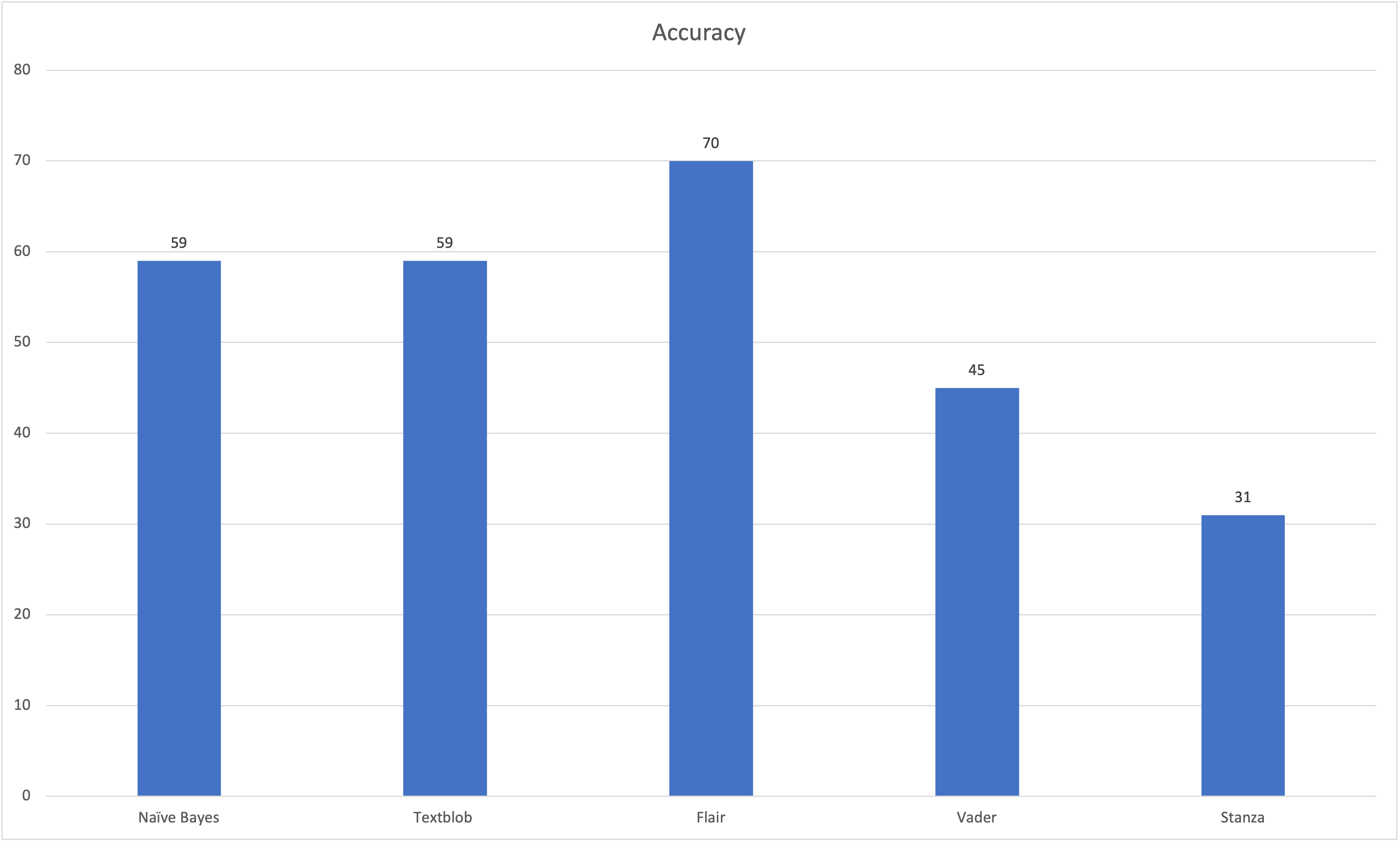}
    \caption{Accuracy of different models}
    \label{fig5}
\end{figure}

Figure \ref{fig5} illustrates the accuracy achieved by various models on our manually annotated dataset.

\section{Conclusion}
\label{sec:conc}

In this research we employed lexicon-based methods for identifying opinion targets in unstructured reviews. By using an unsupervised method, a generalized model that takes an audio file containing an audio data as input and analyses the content identities by translating the audio to text and conducting speaker identification (positive, negative) from the text using different lexicon-based models like Naïve bayes, Textblob, Vader, Stanza and Flair. For testing purposes, we created a dataset consisting of 10,000 manually annotated tweets. The findings showed that Flair outperforms other lexicon-based sentiment analysis techniques by achieving an accuracy of 70\%.

\bibliography{access.bib}{}
\bibliographystyle{IEEEtran}

\end{document}